\documentclass[10pt,twocolumn,letterpaper]{article}

\usepackage{wacv}
\usepackage{times}
\usepackage{epsfig}
\usepackage{graphicx}
\usepackage{amsmath}
\usepackage{amssymb}
\usepackage{subfig}
\usepackage{mathtools}
\usepackage{commath}
\usepackage{adjustbox}
\usepackage{multirow}
\usepackage{float}
\restylefloat{table}
\usepackage{tabularx}

\def\wacvPaperID{87} % Enter the WACV Paper ID here

\wacvfinalcopy % *** Uncomment this line for the final submission
\ifwacvfinal
\usepackage{hyperref}
\hypersetup{breaklinks=true,bookmarks=false}
\else
\usepackage{hyperref}
\hypersetup{pagebackref=true,breaklinks=true,colorlinks,bookmarks=false}

\fi

\ifwacvfinal
\else
\fi

\begin{document}

%%%%%%%%% TITLE
\title{A Vector-based Representation to Enhance Head Pose Estimation}

\author{Zhiwen Cao \thanks{The two first authors made equal contributions.} \and Zongcheng Chu \footnotemark[1]\and Dongfang Liu \and Yingjie Chen\\
Department of Computer Graphics Technology, Purdue University\\
West Lafayette, 47907, USA\\
{\tt\small \{cao270, chu153, liu2538, victorchen\}@purdue.edu}
% \and
% Yingjie Chen

% {\tt\small chu170@purdue.edu}
}

\maketitle

\begin{abstract}

%This paper reveals two issues that exist in current datasets and research works for head pose estimation. 
This paper proposes to use the three vectors in a rotation matrix as the representation in head pose estimation and develops a new neural network based on the characteristic of such representation. We address two potential issues existed in current head pose estimation works:  
1. Public datasets for head pose estimation use either Euler angles or quaternions to annotate data samples. However, both of these annotations have the issue of discontinuity and thus could result in some performance issues in neural network training. 2. Most research works report Mean Absolute Error (MAE) of Euler angles as the measurement of performance. We show that MAE may not reflect the actual behavior especially for the cases of profile views. To solve these two problems, we propose a new annotation method which uses three vectors to describe head poses and a new measurement Mean Absolute Error of Vectors (MAEV) to assess the performance. We also train a new neural network to predict the three vectors with the constraints of orthogonality. Our proposed method  achieves state-of-the-art results on both AFLW2000 and BIWI datasets. Experiments show our vector-based annotation method can effectively reduce prediction errors for large pose angles.
\end{abstract}

\section{Introduction}

%However, most aforementioned studies need to use additional inputs aside from images to conduct the estimation. For instance, some works require supplementary depth information \cite{zakharov2019dpod,xiang2017posecnn,fanelli2011real,liu20163d}, which is usually obtained by depth sensors. Since the depth sensors are not always available, the applications of these methods are limited. Other studies \cite{gu2017dynamic,murphy2008head,li2008apparatus} analyze human head movements from frame sequences by recurrent neural network (RNN)-based approaches. The limitation of this type of research is notable because it can only work under video domain.

Single image head pose estimation is an important task in computer vision which has drawn a lot of research attention in recent years. %A large amount of work has also been done related to face poses such as face alignment \cite{jourabloo2016large}, face landmark detection \cite{sun2013deep,lv2017deep,zhu2012face}, eye gaze estimation \cite{zhang2015appearance,chong2018connecting} and 3D face modeling \cite{jackson2017large,jourabloo2016large,yu2017robust}.  
So far it mainly relies on facial landmark detection \cite{murphy2007head,kumar2017kepler,valle2019face,fanelli2011real}. These approaches show robustness in dealing with scenarios where occlusion may occur by establishing a 2D-3D correspondence matching between images and 3D face models. However, they still have notable limitations when it is difficult to extract key feature points from large poses such as profile views. %This limitation causes significant errors when predicting actual poses. 
To solve this issue, a large array of research has been directed to employ Convolutional Neural Network (CNN) based methods to predict head pose directly from a single image. Several public benchmark datasets \cite{koestinger11a,yang2016wider,zhu2016face,gourier2006head} have been contributed in this area for the purpose of validating the effectiveness of these approaches. Among these approaches, \cite{ruiz2018fine,hsu2018quatnet,yang2019fsa,raytchev2004head} try to address the problem by direct regression of either three Euler angles or quaternions from images using CNN models. %They achieve results with impressive accuracy and lower the error down to a satisfactory level on these public datasets.

However, these studies use either Euler angles or quaternions as their 3D rotation representations. Both Euler angles and quaternions have limitations when they are used to represent rotations. For example, when using Euler angles, the rotation order must be defined in advance. Specifically, when two rotating axes become parallel, one degree of freedom will be lost. This causes the ambiguity problem known as gimbal lock \cite{fua2005monocular}. A quaternion ($\mathbf{q} \in \mathbb{R}^4, ||\mathbf{q}||_2 = 1$)  has the antipodal problem which results in $\mathbf{q}$ and $\mathbf{-q}$ corresponding to the same rotation \cite{saxena2009learning}. In addition, the results from \cite{zhou2019continuity} show that any representation of rotation with four or fewer dimensions is discontinuous. These findings indicate that it is inappropriate to use Euler angles or quaternions to annotate head poses.

\begin{figure}[t]

   \includegraphics[width=\linewidth]{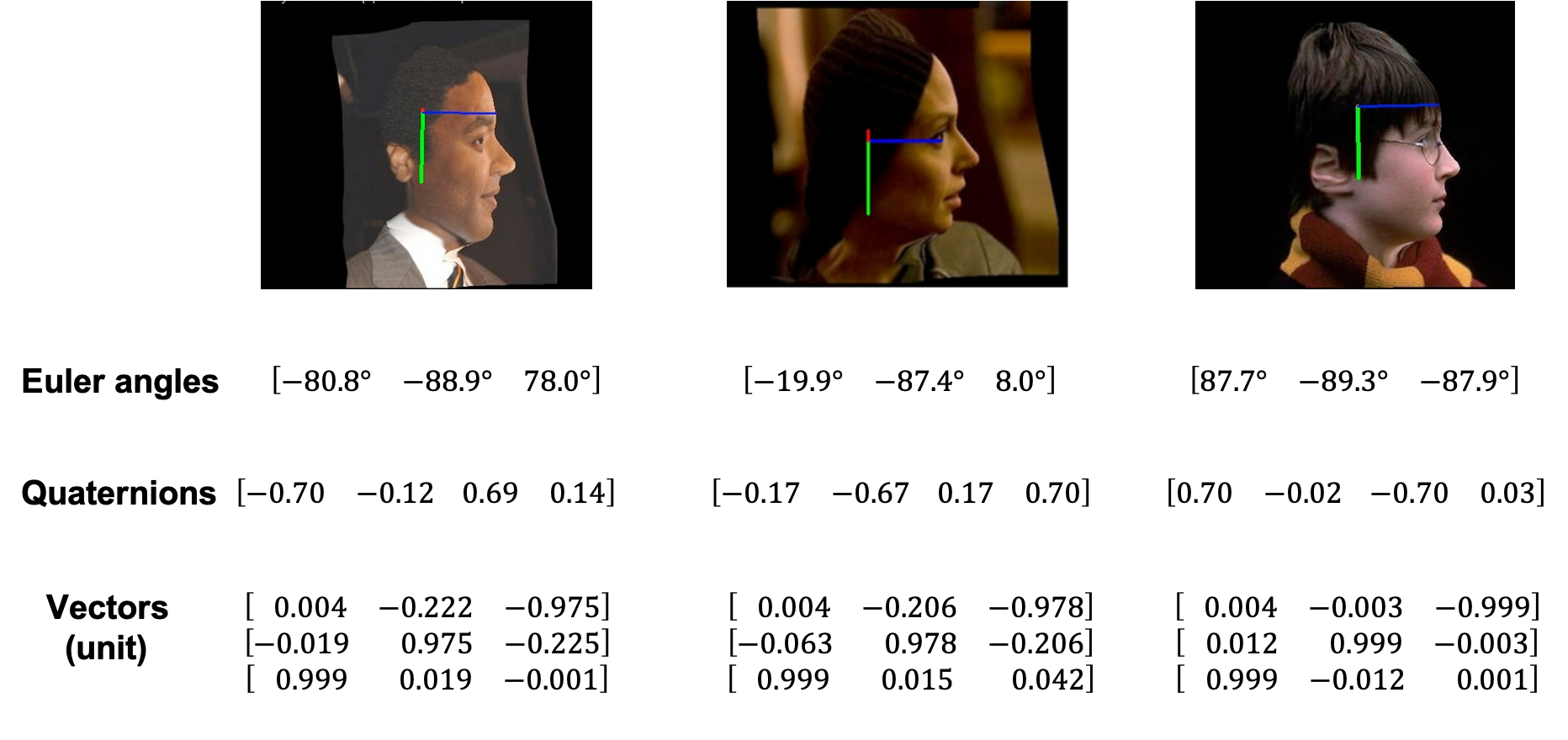}

   \caption{Data samples from 300W-LP dataset and their Euler angles, converted quaternions and three-vector annotations. From top to bottom, three vectors are left (red), down (green) and front (blue)  vectors respectively.}
\label{fig:compare}
\end{figure}

This issue can be illustrated by several samples. Fig.~\ref{fig:compare} shows three images with similar large pose angles from the 300W-LP dataset. However, neither the Euler angle nor quaternion annotations between any two of these show similarity. This leads to two major problems:

(1) It makes training a neural network difficult since the network learns to regress different outputs from the same visual patterns.

(2) It makes mean absolute error (MAE) of Euler angles a problematic measurement of performance. %Most previous works report their results by calculating the MAE of Euler angles between their predictions and the ground truth. 
If the second image is included in the training samples while the first image is the testing case, 
%If a neural network is trained with the second image and then tested on the first image, 
the network's Euler angle prediction will be close to $[-19.9^\circ, -87.4^\circ, 8.0^\circ]$ since it learns an image-to-pose relationship from the second one. 
%It is a good prediction because the first and second image are very similar. 
However, if we compare this prediction with the ground truth $[-80.8^\circ, -88.9^\circ, 78.0^\circ]$,  the MAE will give the result of $44.2^\circ$. This is a large error which cannot reflect the actual model performance. 

% (2) The common Euler angle MAE(mean average error) evaluation metric used in many previous work fails, in that more than one Euler angles representation may result in a same pose. Even an accurate prediction, 

% \textcolor{red}{You need to change the tone here - first say how 3-vectors can solve above problem, then your annotation approach...}

\begin{figure}[t]
\begin{center}
   \includegraphics[width=\linewidth]{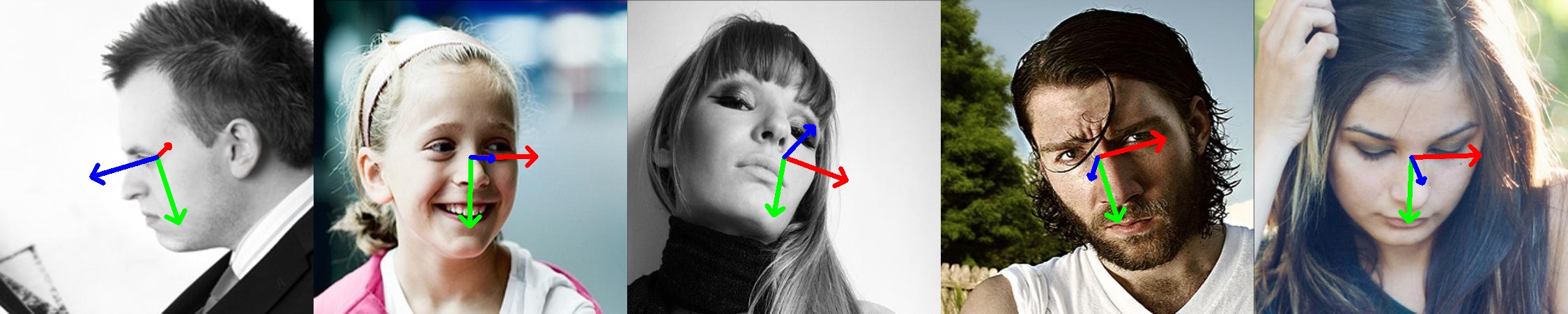}
\end{center}
   \caption{Sample results of head pose estimation by using proposed method.}
\label{fig:demo}
\end{figure}

These issues can be solved by the introduction of three pose vectors. As shown in Fig.~\ref{fig:demo}, head pose can be depicted by a left vector (red), a down vector (green) and a front vector (blue). Using vectors to represent head pose has the following advantages:

(1) It makes the annotations consistent, as we show in the third row in Fig.~\ref{fig:compare}. The vector representations of three images are close to each other. 

(2) Instead of using MAE of Euler angles, we put forward a new measurement which calculates MAE of angles between the vectors (MAEV) that our model predicts and corresponding ground truth. Continue the example above, if the network learns to predict vectors, the MAEV is ${(2.663^\circ + 1.504^\circ + 2.579^\circ)}/{3} = 2.249^\circ$. This is an accurate measure of performance.

\textbf{Specifically, our work contributes at:}

(1) We illustrate that Euler angle annotation has issues of discontinuity and that Euler angle based MAE cannot fully measure the actual performance especially for face profile image.

(2) We instead present MAEV metric that measures angles between vectors, which is a more reliable indicator for the evaluation of pose estimation results.

(3) Based on the vector representation and MAEV, we proposed a deep network pipeline with vectors' orthogonal constraints. To our knowledge, this is the first attempt to formulate head pose estimation problem with vector representation meanwhile consider the vectors' orthogonal constraints in the deep network pipeline.

% In order to leverage the network performance and train different modules jointly, we formulate a multi-loss function.

% Extensive experiment results show that our proposed method outperforms other image based state-of-the-art pose estimation methods on AFLW2000 and BIWI datasets. Additionally, our proposed method shows more consistent prediction error distribution particularly for large degree poses.

%The contributions of our work include: 

%\begin{enumerate}
%    \item We put forward a new vector-based method to represent rotations, which avoids the discontinuity problem of Euler angles and quaternions.
%    \item We propose a new fine-grained CNN model with multi-loss followed by refinements to predict the three vectors.
%    \item We achieve state-of-the-art performances on the AFLW2000, BIWI and AFW benchmark datasets.
%\end{enumerate}

%-------------------------------------------------------------------------
\section{Related Work}

\subsection{Landmark-Based Approaches}

These approaches typically detect key landmarks from images first and then estimate the poses by solving the correspondences between 2d and 3d feature points. \cite{dollar2010cascaded} proposes an algorithm called Cascaded Pose Regression (CPR) which progressively refines a rough initial guess by different regressors in each refinement. \cite{cao2014face} learns a vectorial regression function from training data that it uses to obtain a set of facial landmarks from the image with this function. %\cite{lee2015face} improves the results by the introduction of cascade Gaussian Process Regression Trees (cGPRT) which uses a set of trees to measure the level of similarities of two inputs. 

With the advent of CNN, numerous CNN-based methods have been designed and achieve superior performances than their predecessors. \cite{sun2013deep} puts forward an approach which draws on a three-level convolutional network to estimate the positions of facial landmarks. \cite{zhu2016face} proposes a new cascaded neural network called 3D Dense Face Alignment (3DDFA) which fits a dense morphable 3D face model to the image. They also propose a method to synthesize large-scale training samples in profile views for data labeling. Based on 300W dataset \cite{sagonas2013300}, they create the synthesized 300W-LP dataset which includes 122,450 samples. This has become a widely accepted benchmark dataset. \cite{guo2020towards} makes a step further by proposing a new optimization strategy to regress 3DMM parameters. Their network model simply predicts 9 elements and constructs a rotation matrix from them. As a result, this can never guarantee it to be a rotation matrix.  %\cite{bulat2017far} builds a very large dataset named LS3D-W for 3d face alignment which unifies all the existing datasets. It also constructs two very powerful networks to perform 2d and 3d face alignments and both networks achieve impressive results.

Some methods treat head pose estimation as an auxiliary task. They perform various facial related tasks jointly with CNN. \cite{ranjan2017hyperface} proposes Hyperface which uses a single CNN model to perform face detection, pose estimation, feature localization and gender recognition simultaneously. % \cite{ranjan2017all} goes a step further. They add more tasks on one CNN model including age estimation, face recognition and smile detection.
\cite{kumar2017kepler} proposes a H-CNN (Heatmap-CNN) which refines the locations of the facial keypoints iteratively and provides the pose information in Euler angles as a by-product.

These methods rely heavily on the quality of landmark detection. If it fails to detect the landmark accurately, a large error will be introduced.%Furthermore, recent landmark-free approaches show that performing pose estimation does not necessarily need landmark detection.

% \subsection{CNN-based Methods}
% \indent Recent work using deep neural networks has achieved great success. Such work deploys CNNs to learn an end-to-end mapping from a single RGB or RGB-D image to the actual object poses. Compared with the traditional methods mentioned above, deep neural networks (DNNs) are more robust against the changes of the environment. A myriad of researches have been conducted in the past few years since advances in deep learning (such as the GPU-support computing and open-source framework) make it possible to easily train complex CNNs on large datasets.

\subsection{Landmark-Free Approaches}
The latest state-of-the-art landmark-free approaches explore the research boundary and improve the results by a significant margin. \cite{ruiz2018fine} puts forward a CNN combined with multi-losses. It predicts three Euler angles directly from a single image and outperforms all the prior landmark based methods. \cite{hsu2018quatnet} further presents their quaternion based approach which avoids the gimbal lock issue lying in Euler angles. \cite{yang2019fsa} proposes a CNN model using a stage-wise regression mechanism. They also adopt an attention mechanism combined with a feature aggregation module to group global spacial features. \cite{liu2019facial} treats pose estimation as a label distribution learning problem. They associate a Gaussian label distribution instead of a single label with each image and train a network which is similar to Hopenet \cite{ruiz2018fine}.

% \subsection{Representation of Rotation}
% There are a series of ways to represent a rotation in a 3D world. Euler angle \cite{wiki:eulerangle}, quaternion \cite{wiki:quaternion}, axis-angle \cite{wiki:axis_angle} and lie algebra \cite{wiki:liealgebra}. They describe the rotation in a compact form with at most 4 dimensions. However, \cite{zhou2019continuity} shows that it needs at least 5 dimensions of information to achieve a continuous representation of rotations in 3D space which means all the above representation methods will have the same issue of discontinuity as demonstrated in section 1. This makes rotation matrix \cite{wiki:rotation_matrix} a good alternative for this task. A 3d rotation matrix has 9 elements and can be described as orthogonal matrices with determinant equals to $+1$. The set of all the rotation matrices forms a continuous special orthogonal group $SO(3)$. When it is used to describe rotation, it doe not have problem of discontinuity or ambiguity. This inspires us to use three orthogonal vectors to represent rotations. In section 3.1, we show that predicting three head pose vectors is in fact equivalent to predicting the rotation matrices.

\subsection{6D Object Pose Estimation}

6D object pose estimation from RGB images includes estimation of 3D orientation and 3D location. The task of orientation estimation resembles our head pose estimation one. The approaches can be divided into two categories: \cite{peng2019pvnet, zakharov2019dpod, song2020hybridpose} first estimate the object mask to determine  its location in the image, then build the correspondence between the image pixels and the available 3D models. After that, The 6d pose can be solved through PnP algorithm \cite{lepetit2009epnp}. The other type of methods such as \cite{xiang2017posecnn, mahendran20173d, gao2018occlusion} use network to predict orientation directly. However, they use either axis-angle or quaternion as their representations of rotation and none of them notice the problem of discontinuity.

%-----------------------------------------------------------------
\section{Method}

In this section, we first present a thorough discussion on our vector-based representation and how we formulate the problem (Sec. 3.1). Then, we give an overview of the our network structure (Sec. 3.2). Prediction module implementation is described in Section 3.3. A multi-loss training strategy is then introduced in Section 3.4. Finally, by means of Singular Value Decomposition (SVD), we obtain three orthonormal vectors (Sec. 3.5).

\subsection{Representation of Rotation}

There are various ways to represent a rotation in a 3D world. Euler angle, quaternion, axis-angle and lie algebra. They describe the rotation in a compact form with at most 4 dimensions. However, \cite{zhou2019continuity} shows that it needs at least 5 dimensions of information to achieve a continuous representation of rotations in 3D space which means all the above representation methods will have the same issue of ambiguity as demonstrated in section 1. This makes rotation matrix a good alternative. A 3d rotation matrix has 9 elements and can be described as orthogonal matrices with determinant equals to $+1$. The set of all the rotation matrices forms a continuous special orthogonal group $SO(3)$. When it is used to describe rotation, it doe not have problem of discontinuity or ambiguity.

The question left is what metric we should adopt to measure the closeness of two rotation matrices. A straightforward way is to measure the Frobenius norm of two rotation matrices, i.e. the square root of the sum of squares of differences of all $9$ elements. If we define the left, down and front vectors at the reference starting point to be $\boldsymbol{v}_1=\begin{bmatrix} 1,  0,  0 \end{bmatrix}^T$, $\boldsymbol{v}_2=\begin{bmatrix} 0,  1,  0 \end{bmatrix}^T$ and $\boldsymbol{v}_3=\begin{bmatrix} 0,  0,  1 \end{bmatrix}^T$ respectively. After applying a rotation matrix $\boldsymbol{R}_{3\times 3} = \begin{bmatrix} \boldsymbol{r}_1,  \boldsymbol{r}_2,  \boldsymbol{r}_3 \end{bmatrix}$ where $\boldsymbol{r}_i$ denotes the $i^{th}$ column vector in $\boldsymbol{R}$, the three vectors then become $\boldsymbol{v}'_1 = \boldsymbol{R}\boldsymbol{v_1} = \boldsymbol{r}_1$, $\boldsymbol{v}'_2 = \boldsymbol{R}\boldsymbol{v_2} = \boldsymbol{r}_2$ and $\boldsymbol{v}'_3 = \boldsymbol{R}\boldsymbol{v_3} = \boldsymbol{r}_3$. The equations show that three vectors of head pose is in essence equivalent to the three columns of rotation matrices. As a result, Frobenius norm is equivalent to $\sqrt{d_1^2 + d_2^2 + d_3^2}$ in Fig.~\ref{fig:vec_refine}.

Even though Frobenius norm is an accurate measurement, it is hard for we human beings to perceive the difference of rotation angles through the distance between endpoints of pose vectors. Therefore, we put forward a new metric which is more intuitive: the mean absolute error of vectors (MAEV). For each vector, we compute absolute error between the ground truth and predicted one, then we obtain MAEV by calculating the mean value of three errors.

The problem of head pose estimation thus can be defined as: given a set of $N$ training images $\boldsymbol{X} = \{x^{(1)}, x^{(2)}, \cdots, x^{(N)}\}$, find a mapping function $F$ such that estimates $\boldsymbol{\hat{R}}^{(i)}=F(x^{(i)})$ where $\boldsymbol{\hat{R}}^{(i)} = \begin{bmatrix}\boldsymbol{\hat{r}_1}^{(i)}, \boldsymbol{\hat{r}_2}^{(i)}, \boldsymbol{\hat{r}_3}^{(i)} \end{bmatrix}$ that matches the ground truth rotation matrix $\boldsymbol{R}$ as close as possible. We try to find an optimal $F$ for all $\boldsymbol{X}$ by minimizing the sum of squared $L_2$ norm between the predicted and ground truth vectors.

\begin{align}
    \resizebox{.9\hsize}{!}{$\mathcal{L} = \frac{1}{N}\sum_{i=1}^{N} \norm[2]{\boldsymbol{r_{1}^{(i)}} - \boldsymbol{\hat{r}_{1}^{(i)}}}_2^2 +  \norm[2]{\boldsymbol{r_{2}^{(i)}} - \boldsymbol{\hat{r}_{2}^{(i)}}}_2^2 + \norm[2]{\boldsymbol{r_{3}^{(i)}} - \boldsymbol{\hat{r}_{3}^{(i)}}}_2^2$}
\end{align}

% Since the special property of a rotation matrix $\mathbf{R}_{3\times 3}$ of having three orthogonal column vectors, and easy conversion between Euler angles and its corresponding rotation matrix. We decide to adopt such column vectors as our vector based representation. Because rotation matrices form a continuous special orthogonal group $SO(3)$, and have a non-ambiguous representation \cite{saxena2009learning}. Using three vectors to describe rotations shares the same advantages and avoid the gimbal lock and antipodal problem otherwise would incur.

\subsection{TriNet Overview}

\begin{figure*}[!htbp]
\centering
\includegraphics[width=0.8\textwidth]{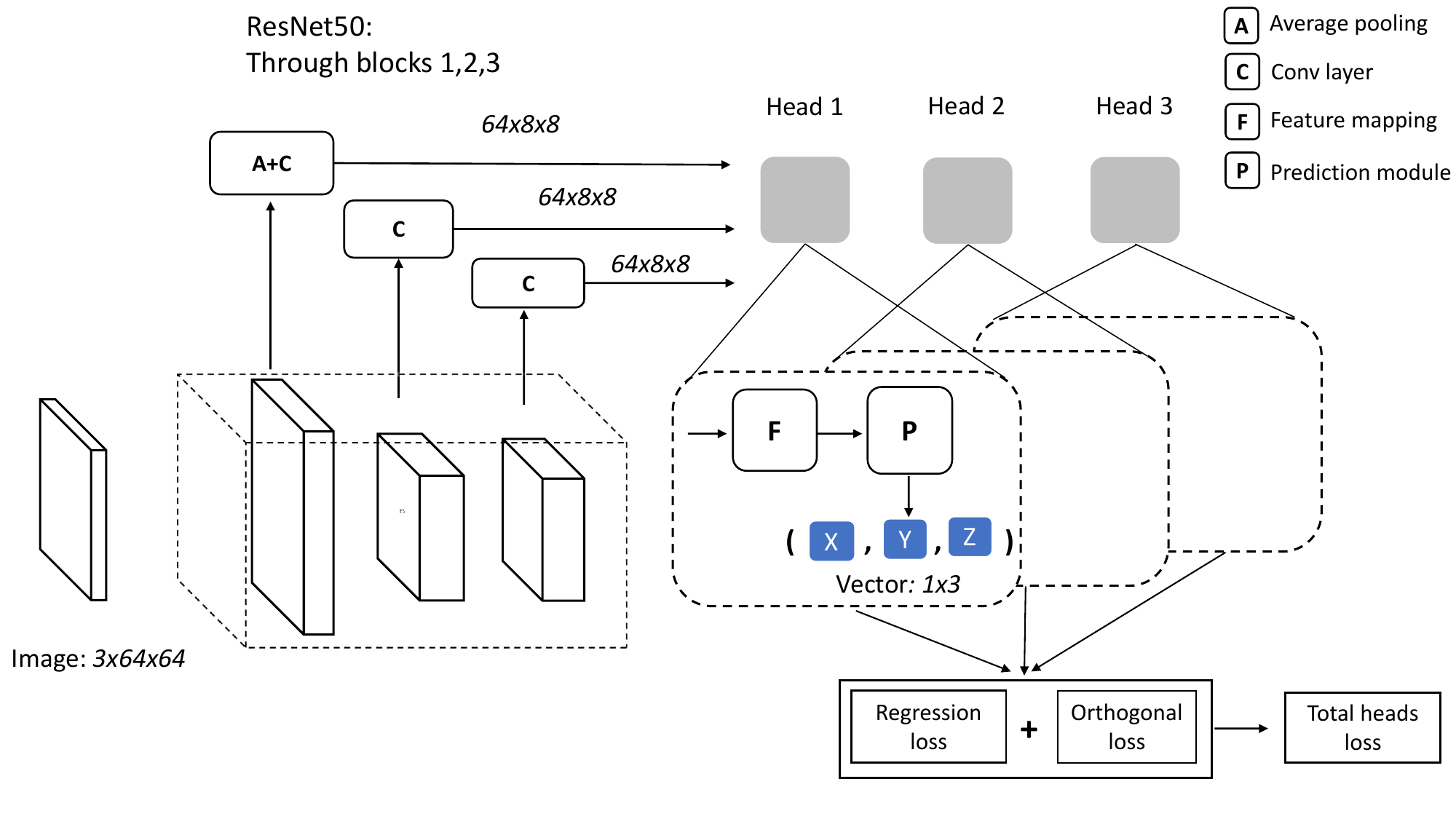}
\caption{Overview of the network.}
\label{fig:overview}
\end{figure*}

\begin{figure}[!htbp]
    \begin{center}
    \includegraphics[width=0.8\columnwidth]{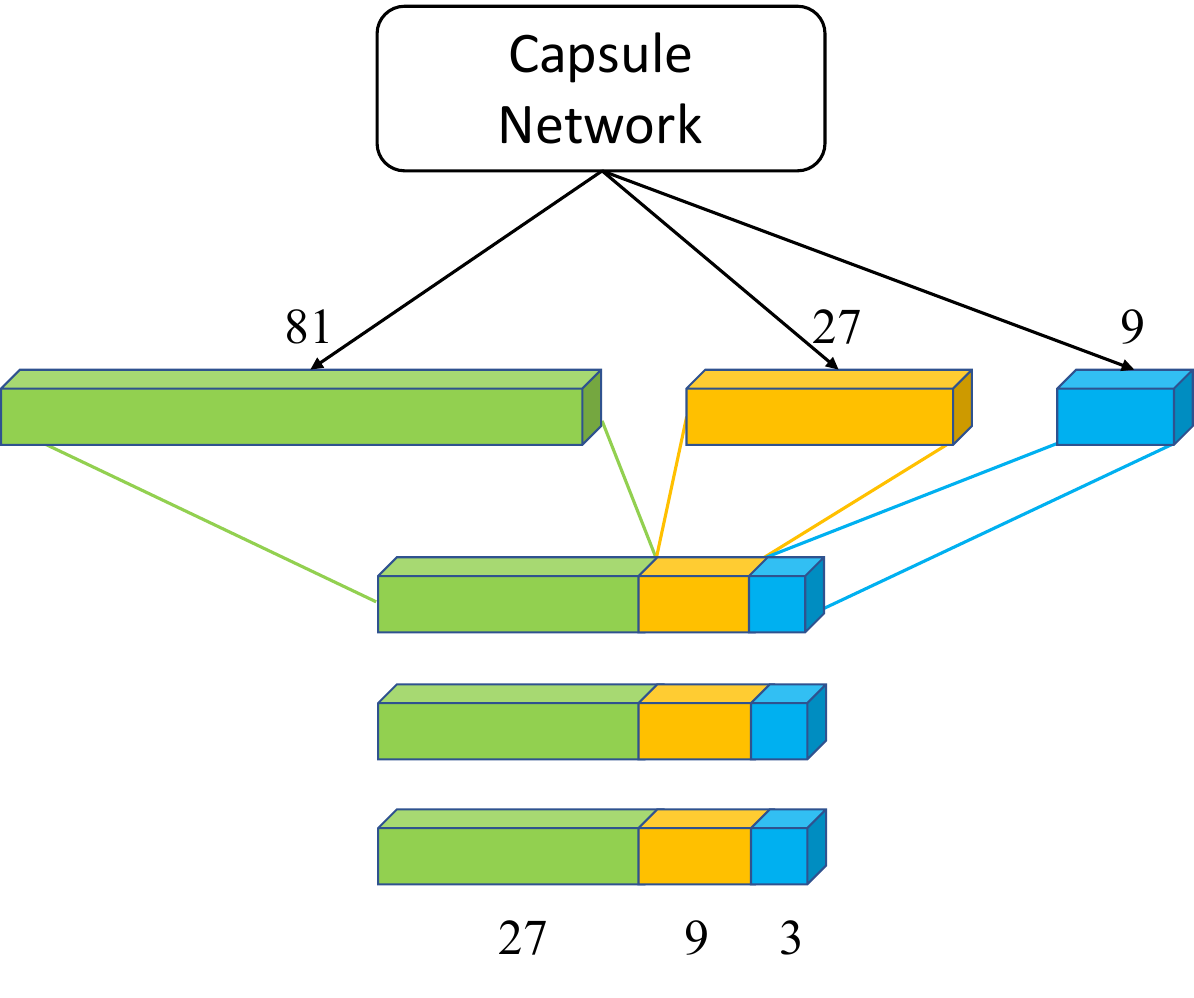}
    \end{center}
    \caption{Details of the prediction module.}
    \label{fig:prediction}
    
\end{figure}

%---------- Connection with vector representation ----------%

Rotation matrix is 9-D dimensional representation which requires the network to predict 9 elements. There is no off-the-shelf network model that we can adopt to perform this task, so we design our TriNet shown as Fig.~\ref{fig:overview}. TriNet is composed of one backbone and three head branches. Each head follows the coarse-to-fine strategy, constitutes a feature mapping and prediction module and is responsible for predicting one vector alone. 
Ideally, three vectors should be perpendicular to each other, so we further introduce an orthogonal loss function which punishes the model if the predicted ones are not orthogonal.

%---------- Connection with vector representation ----------%

An input image with fixed size goes through a backbone network (ResNet50 in Fig.~\ref{fig:overview}). We define $S$ stages and at each stage $s$, a feature map is extracted from the output of an intermediate layer of the backbone network. These are considered as candidate features and fed into the feature grouping module. For feature grouping component, we follow the same implementation as FSA-Net \cite{yang2019fsa}. %  Capsule network can learn and build a part-to-whole hierarchical relationship between low level and higher level features and attention mechanism can learn the local and global spatial importance of input features.}
Since the grouping module requires uniform shape ($w\times h \times c$) of input features, we apply average pooling to reduce the feature map size to $w\times h$ and use $c$ $1\times 1$ convolution operations to transform the feature channels into $c$. The feature grouping module outputs $3$ $c'$-dimensional vectors.

We then feed them to the prediction module to regress one pose vector. Since three head branches share the identical structure, the other two pose vectors can be obtained in the same way by going through different head branches.

\subsection{Prediction Module}
The prediction module follows the strategy of coarse-to-fine multi-stage regression. Features extracted from shallow layers are responsible for performing coarse predictions. As the network goes deeper, the high level features become more informative and can be used for fine-grained and more accurate predictions. Since each component of a unit vector is within the range of $[-1,1]$, for each stage, we divide the range into different numbers of intervals. The deeper the layer is, the more intervals the range $[-1, 1]$ will be divided into. The prediction module performs the estimation by taking the average of the expectation values from all $S$ stages together:

\begin{align}
    \hat{y} &= \frac{1}{S} \sum_{s=1}^{S}  \sum_{i=1}^{n^{(s)}} p_{i}^{(s)} \cdot q_{i}^{(s)}
\end{align}

where $n^{(s)}$ is the number of intervals at stage $s$, $p_{i}^{(s)}$ is the probability that the element is in the $i^{th}$ interval and $q_{i}^{(s)}$ is the mean value of the $i^{th}$ interval.

\subsection{Training Objective}
The training objective involves multiple losses: regression loss $\mathcal{L}_{reg}$ and the orthogonal loss $\mathcal{L}_{ortho}$ which measures the orthogonality between each pair of the predicted vectors. The overall objective loss is the weighted sum of two losses:

\begin{align}
    \mathcal{L} &= \mathcal{L}_{reg}(\boldsymbol{v}_i,\hat{\boldsymbol{v}}_i) + \alpha \mathcal{L}_{ortho}(\hat{\boldsymbol{v}}_i,\hat{\boldsymbol{v}}_j)
\end{align}

where $\hat{\boldsymbol{v}}_i$ and $\boldsymbol{v}_i$ are the $i^{th}$ predicted and ground truth vectors respectively. The weighted term $\alpha$ is set to a small number whose range is between $[0.1,0.5]$. It best setting is found through experiments. Each loss term is shown as follows:

\begin{align}
        \mathcal{L}_{reg} &= \sum_{i=1}^{3} \text{mse}(\boldsymbol{v}_i, \hat{\boldsymbol{v}}_{i})\\
    \mathcal{L}_{ortho} &= \sum_{i \neq j}\text{mse}(\hat{\boldsymbol{v}}_i \hat{\boldsymbol{v}}_j, 0) \text{ where } i,j = 1,2,3\\
    \nonumber
\end{align}

We adopt mean square error loss function for both regression loss and orthogonal loss.

\subsection{Vector Refinement}
Even though we impose orthogonal constraints $\mathcal{L}_{ortho}$ in the loss function, the three vectors that TriNet predicts are still not perpendicular to each other. Therefore, it is necessary to select three orthogonal vectors to match the predicted vectors as close as possible. 
% Even though we impose a penalty term $\mathcal{L}_{ortho}$ in the loss function as orthogonality constraints between each pair of vectors, the three vectors that TriNet predicts may still not be perpendicular to each other. Therefore, it’s necessary to select the three orthogonal vectors to match the estimated vectors as close as possible. 

This problem can be stated as: Given a noisy predicted matrix $\boldsymbol{R}$, find the closest rotation matrix $\boldsymbol{\hat{R}}$ and the measure of closeness needs to have physical meaning. A naive way to find a rotation matrix from a noisy matrix is applying the Gram–Schmidt process to either its rows or columns. Its simple geometric interpretation makes it very popular, however, the result is rather arbitrary because this method depends on which two rows or columns of $\boldsymbol{R}$ are selected.

This paper adopts the measure of Euclidean or Frobenius norm of $\boldsymbol{R} - \boldsymbol{\hat{R}}$. % In other words, we need to minimize the square root of the sum of squares of all $9$ elements in $\boldsymbol{R} - \boldsymbol{\hat{R}}$. 
It can be expressed by the following formula:

\begin{align}
\text{min }& || \boldsymbol{\hat{R}}  - \boldsymbol{R}||_F \nonumber\\
\text{ subject to } & \boldsymbol{\hat{R}}^T \boldsymbol{\hat{R}} = \boldsymbol{I} \text{ and det} {\boldsymbol{\hat{R}}}=+1
\end{align} 

The reasons for choosing Frobenius norm are as follows:

\begin{figure}[t]
    \begin{center}
    \includegraphics[clip,width=40mm]{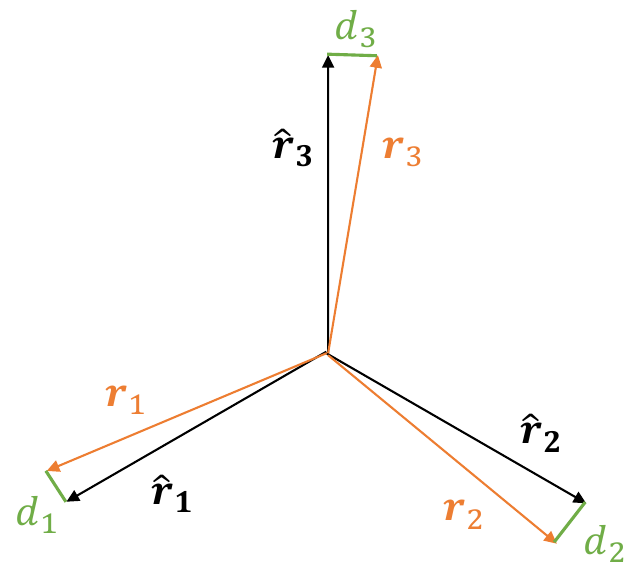}
    \end{center}
    \caption{ Given a matrix $\boldsymbol{R} = \begin{pmatrix} \boldsymbol{r}_1,
    \boldsymbol{r}_2,\boldsymbol{r}_3\end{pmatrix}$, find the closest rotation matrix $\boldsymbol{\hat{R}} = \begin{pmatrix} \boldsymbol{\hat{r}}_1,
    \boldsymbol{\hat{r}}_2,\boldsymbol{\hat{r}}_3\end{pmatrix}$. The Frobenius norm of the difference between $\boldsymbol{R}$ and $\boldsymbol{\hat{R}}$ is equal to $\sqrt{d_1^2 + d_2^2 + d_3^2}$}
    \label{fig:vec_refine}
    
\end{figure}

(1) It has a simple geometric interpretation (see Fig.~\ref{fig:vec_refine}).

(2) The solution is unique and can be obtained by a closed-form formula \cite{sarabandi2020closed}.

\cite{mao1986optimal} shows that given a matrix $\boldsymbol{R} = \boldsymbol{U} \boldsymbol{\Sigma} \boldsymbol{V}^T$, the optimal solution can be achieved by $\hat{\boldsymbol{R}} = \boldsymbol{U} \boldsymbol{V}^T$. This method does not guarantee that $\det(\boldsymbol{\hat{R}}) = +1$. If a highly noisy matrix $\boldsymbol{R}$ is given, $\det(\boldsymbol{\hat{R}}) = -1$ may happen. If this is the case, the closest rotation matrix can be obtained by 

\begin{equation}
    \boldsymbol{\hat{R}} = \boldsymbol{U}\text{diag}(1, 1, -1)\boldsymbol{V}^T
\end{equation}

% \subsection{Data Augmentation Strategy}

% \begin{figure}[t]
% \includegraphics[width=1\columnwidth]{imgs/augmentation.png}
% \caption{Visualization of different types of data augmentation strategies. From left to right: original, random crop with white padding, random crop with black padding, random zoom and cutout.}
% \label{fig:augmentation}
% \end{figure}

% We borrow the data augmentation strategies from \cite{yang2019fsa} and apply uniformly on the competing methods. Fig.~\ref{fig:augmentation} shows different types of augmentation strategies we use. Although data augmentation for supervised image classification has been studied extensively, few efforts have been made to the domain of pose estimation. We mainly apply the crop technique with random zoom to augment the original images. We additionally add white and black paddings to generate cropped images with more variety. Besides, cutout is adopted by masking out two random locations of an image. %of a whole image.

% To prevent a trivial solution when applied exclusively on the entire head portion, cutout is used by 

\section{Experiments}

\begin{table*}[!htbp]
\centering
\caption{Mean absolute errors of Euler angles and vectors on AFLW2000. All trained on 300W-LP. Values in () are converted from the other side. Methods with $*$ are not open source. Their results are claims from authors. }
\label{tab:aflw2000}
\begin{tabular}{c|cccc|cccc}
\hline
\multirow{2}{*}{Method} & \multicolumn{4}{c|}{Euler angles errors}                                                                    & \multicolumn{4}{c}{Vector errors}                                                                         \\ \cline{2-9} 
                        & \multicolumn{1}{c}{roll}   & \multicolumn{1}{c}{pitch} & \multicolumn{1}{c}{yaw} & \multicolumn{1}{c|}{MAE} & \multicolumn{1}{c}{left} & \multicolumn{1}{c}{down} & \multicolumn{1}{c}{front} & \multicolumn{1}{c}{MAEV} \\ \hline
3DDFA\cite{zhu2016face}                 & \multicolumn{1}{c}{28.432} & 27.085                    & 4.710                   & 20.076                   & (30.570                   & 39.054                   & 18.517                    & 29.380)                   \\
Dlib\cite{kazemi2014one}                   & 22.829                     & 11.250                    & 8.494 & 14.191 & (26.559 & 28.511 & 14.311 & 23.127) \\
Hopenet\cite{ruiz2018fine}                 & 6.132                      & 7.120                     & 5.312                   & 6.188                    & (7.073 & 5.978 & 7.502 & 6.851) \\
FSA-Net\cite{yang2019fsa} & 4.776 & 6.341 & 4.963 & 5.360 & (6.753 & 6.215 & 7.345 & 6.771) \\
Quatnet\cite{hsu2018quatnet}$^{*}$                 & \textbf{3.920}                      & \textbf{5.615}                     & \textbf{3.973}                   & \textbf{4.503}                    & -                        & -                        & -                         & -                        \\
HPE\cite{huang2020improving} $^{*}$                   & 4.800                      & 6.180                     & 4.870                   & 5.280                    & -                        & -                        & -                         & -                        \\
TriNet & (4.042 & 5.767 & 4.198 & 4.669) & \textbf{5.782} & \textbf{5.666} & \textbf{6.519} & \textbf{5.989} \\ \hline
\end{tabular}
\end{table*}

\begin{table*}[!htbp]
\centering
\caption{Mean absolute errors of Euler angles and vectors on BIWI. All trained on 300W-LP. Values in () are converted from the other side. Methods with $*$ are not open source. Their results are claims from authors.}
\label{tab:biwi}
\begin{tabular}{c|cccc|cccc}
\hline
\multirow{2}{*}{Method} & \multicolumn{4}{c|}{Euler angles errors}                                                                    & \multicolumn{4}{c}{Vector errors}                                                                         \\ \cline{2-9} 
                        & \multicolumn{1}{c}{roll}   & \multicolumn{1}{c}{pitch} & \multicolumn{1}{c}{yaw} & \multicolumn{1}{c|}{MAE} & \multicolumn{1}{c}{left} & \multicolumn{1}{c}{down} & \multicolumn{1}{c}{front} & \multicolumn{1}{c}{MAEV} \\ \hline
3DDFA\cite{kazemi2014one}                   & \multicolumn{1}{c}{13.224} & 41.899 & 5.497  & 20.207 & (23.306 & 45.001 & 35.117 & 34.475) \\
Dlib\cite{kazemi2014one} & 19.564 & 12.996 & 11.864 & 14.808 & (24.842 & 21.702 & 14.301 & 20.282) \\
Hopenet\cite{ruiz2018fine} & 3.719 & 5.885 & 6.007 & 5.204 & (7.650 & 6.728 & 8.681 & 7.687) \\
FSA-Net\cite{yang2019fsa}                 & 3.069 & 5.209 & 4.560 & 4.280 & (6.033 & 5.959  & 7.218 & 6.403) \\
Quatnet\cite{hsu2018quatnet}$^{*}$ & \textbf{2.936} & 5.492 & 4.010 & 4.146 & -& - & -  & -  \\
HPE\cite{huang2020improving} $^{*}$ & 3.120 & 5.180 & 4.570 & 4.290 & - & -  & - & - \\
TriNet & (4.112 & \textbf{4.758} & \textbf{3.046}  & \textbf{3.972}) & \textbf{5.565}                    & \textbf{5.457}                    & \textbf{6.571}                     & \textbf{5.864}                    \\ \hline
\end{tabular}
\end{table*}

\begin{table*}[!htbp]
\centering
\caption{Mean absolute errors of Euler angles and vectors on BIWI. 70\% of the data is used for training and the remaining 30\% is for testing. Values in () are converted from the other side.}
\label{tab:biwi3070}
\begin{tabular}{c|cccc|cccc}
\hline
\multirow{2}{*}{Method} & \multicolumn{4}{c|}{Euler angles} & \multicolumn{4}{c}{Vectors} \\ \cline{2-9} 
                        & roll   & pitch   & yaw    & MAE   & left  & down  & front  & MAEV \\ \hline

Hopenet\cite{ruiz2018fine} & 4.334 & 4.420 & 4.094 & 4.283 & (6.465 & 6.272 & 6.268 & 6.335) \\
FSA-Net\cite{yang2019fsa} & 4.056 & 4.558 & 3.155  & 3.922 & (5.854 & 6.189 & 5.440 & 5.828) \\
TriNet & (\textbf{2.928} & \textbf{3.035} & \textbf{2.440} & \textbf{2.801}) & \textbf{4.067} & \textbf{4.140} & \textbf{3.976} & \textbf{4.061} \\ \hline
\end{tabular}
\end{table*}

\begin{figure*}
\centering
\includegraphics[width=0.8\textwidth]{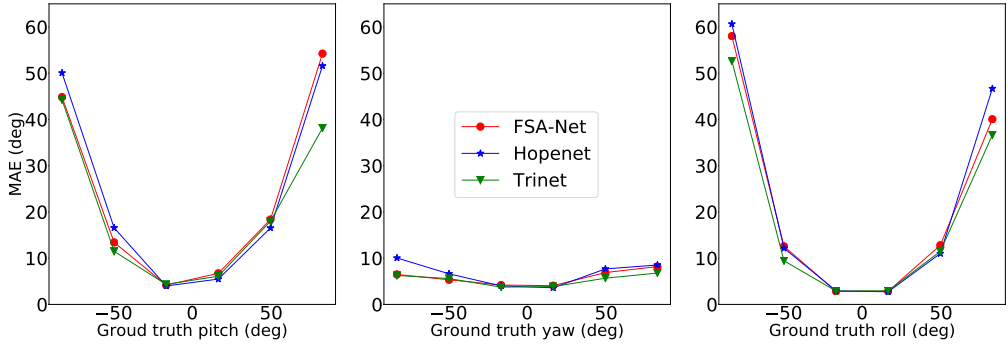}
\caption{MAE on AFLW2000 using landmark-free methods. All trained on 300W-LP.}
\label{fig:eulererroraflw2000}
\end{figure*}

\begin{figure*}
\centering
\includegraphics[width=0.8\textwidth]{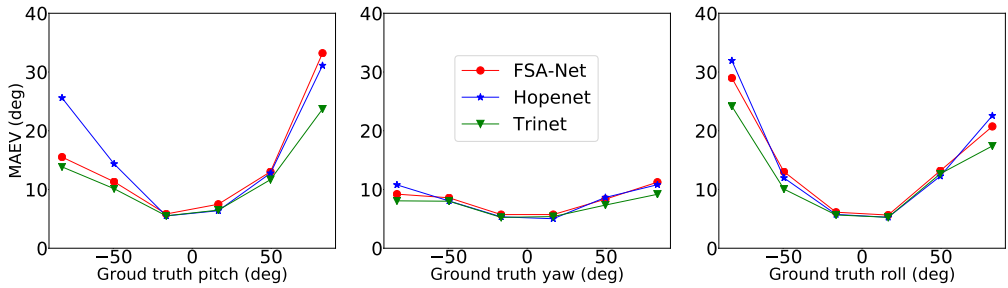}
\caption{MAEV on AFLW2000 using landmark-free methods.  All trained on 300W-LP.}
\label{fig:vectorerroraflw2000}
\end{figure*}

\begin{figure}
\centering
\includegraphics[width=\columnwidth]{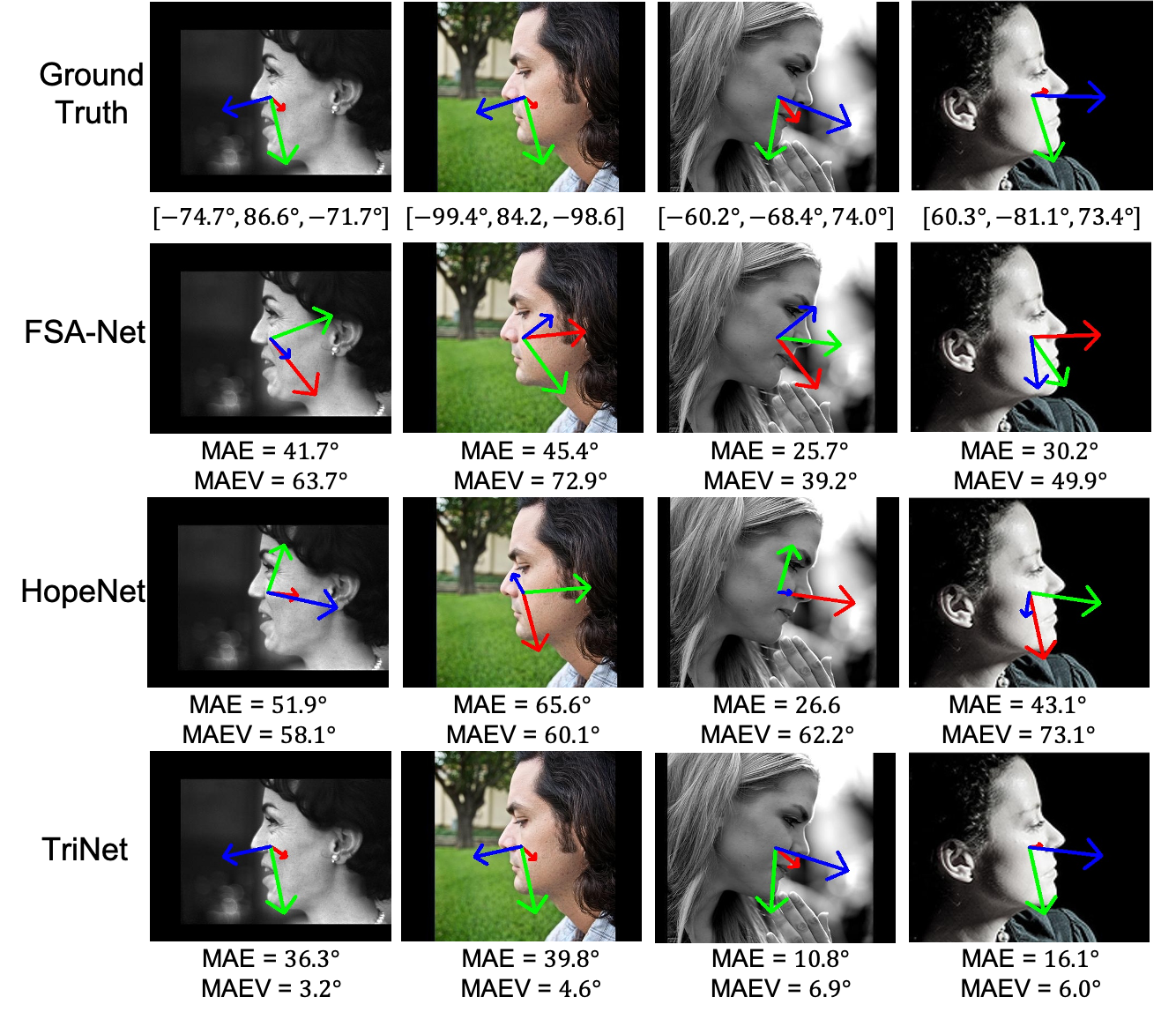}
\caption{Comparison of pose estimation results on AFLW2000 profile images. All trained on 300W-LP.}
\label{fig:euler_failure_cases}
\end{figure}

\subsection{Implementation Details}
We  implement  our  proposed  network  using  Pytorch. We follow the data augmentation strategies from \cite{yang2019fsa} and apply uniformly on the competing methods. We  train  the  network  using  Adam  optimizer  with  an initial learning rate of 0.0001 over 90 epochs. The learning rate decay parameter is set to be 0.1 for every 30 epochs. %The experiments are conducted on a lab PC with two RTX 2080 Ti GPU support. %It takes about 3 hours to complete the whole training process on 300W-LP dataset \cite{zhu2016face} with the batch size of 64 and image input size of $64\times 64$.

\subsection{Datasets and Evaluation}
Our experiments are based on three popular public benchmark datasets: 300W-LP \cite{zhu2016face}, AFLW2000 \cite{zhu2015high}, and BIWI \cite{fanelli2013random} datasets.

\textbf{300W-LP}
The 300W-LP dataset \cite{zhu2016face} is expanded from 300W dataset \cite{sagonas2013300} which is composed of several standardized datasets, including AFW \cite{zhu2012face}, HELEN \cite{zhou2013extensive}, IBUG \cite{sagonas2013300} and LFPW \cite{belhumeur2013localizing}. 
By means of face profiling, this dataset generates 122,450 synthesized images based on around 4,000 pictures from the 300W dataset.

\textbf{AFLW2000}
The AFLW2000 \cite{zhu2015high} dataset contains 2,000 images which are the first 2000 images the AFLW dataset \cite{koestinger11a}. 
%The pose information is obtained by fitting a mean 3D model \cite{storer20093d} to the annotated landmarks on the images. 
This dataset possesses a wide range of varieties in facial appearances and background settings %which makes it a good dataset for testing the behavior of our proposed method.

\textbf{BIWI}
The BIWI dataset \cite{fanelli2013random} contains 15,678 pictures of 20 participants in an indoor environment. Since the dataset does not provide bounding boxes of human heads, we use MTCNN \cite{zhang2016joint} to detect human faces and loosely crop the area around the face to obtain face bounding boxes results. 
%In addition, it also provides depth information for each image frame.

% \textbf{AFW} 
% Some research works report the results onAF The AFW dataset \cite{yang2016wider} is a very small dataset that only has 205 images with 468 human faces and their poses are coarsely annotated with the accuracy of only $15^{\circ}$. report the percentage that the difference of yaw angles between prediction and ground truth are less than $15^{\circ}$. This is a very rough evaluation and many works have achieved more than $95\%$ accuracies. Therefore, an evaluation performed on AFW is not meaningful and we dismiss it from our benchmark options.

In order to compare to the state-of-the-art methods, we follow the same training and testing setting as mentioned in Hopenet \cite{ruiz2018fine} and FSA-Net \cite{yang2019fsa}. Notice that we also filter out test samples with Euler angles that are not in the range between $-99^{\circ}$ and $99^{\circ}$ to keep consistent with the strategies used by Hopenet and FSA-Net.  We implement our experiments in two scenarios:

(1) We train the models on 300W-LP and test on two other datasets: AFLW2000 and BIWI. 

(2) We apply a $3$-fold cross validation on BIWI dataset and report the mean validation errors. We split the dataset into $3$ groups and ensure that the images of one person should appear in the same group. Since there are $24$ videos in the BIWI dataset, each group contains 8 videos and in a round we have $16$ videos for training and $8$ for testing. 
%Even though this dataset provides different sources including RGB image, depth and time sequence, our TriNet performs the prediction using only RGB image.

For all the experiments above, we report both the MAE of Euler angles and MAEV as results. 
%We report both MAE of Euler angles and MAEV in our experiment results. We find that a method with lower MAE does not necessarily  perform better than others under the measurement of MAEV. Putting two measurements together can help us tell the difference between them and show the advantages of MAEV.

% However, the discrepancy of Euler angles representation for large poses makes it difficult to reflect the real error distribution of poses near $ \pm 90^{\circ}$. To address this problem, we further formulate MAEV metric for measuring the angle errors between prediction vectors and ground truth vectors. 

\subsection{Comparison to State-of-the-art Methods}

We compare our proposed TriNet with other state-of-the-art methods on public benchmark datasets. To make a fair comparison, we rerun the open-sourced models and ours under the same experiment environment and measure the results by both MAE and MAEV. For those which are not open sourced, we cite their MAE results claimed by the authors in our tables for reference.

Facial landmark based approach 3DDFA \cite{zhu2016face} tries to fit a dense 3D model to an RGB image through a Cascaded CNN architecture. The alignment framework applies to large poses up to $90$ degrees. Hopenet \cite{ruiz2018fine} proposes a fine-grained structure by combining classification loss and regression loss to predict the head pose in a more robust way. Quatnet \cite{hsu2018quatnet} uses quaternions labeling data for training the model to avoid the ambiguity of Euler angle representation. FSA-Net \cite{yang2019fsa} proposes a network which combines a stage-wise regression scheme and a feature grouping module for learning aggregated the spatial features. \cite{huang2020improving} proposes to use two stage method which treats classification and regression separately and averages top-k outputs as pose regression subtask.

% \textcolor{red}{why Quatnet does not have vector value?}
% \textcolor{red}{The table may confuse reviewer. Two methods - one: Euler angles (convert to Vectors, Left/down, front, MAVE), or use () to indicate these values are converted. }

\subsection{Experiment Results}

%---------- Ablation Tables ----------%
\begin{table*}[!htbp]
\centering
\caption{Ablation study for different feature mapping methods (with/ without attention mapping) and loss items (with/ without orthogonality loss) and capsule network (with/without capsule network). Trained on 300W-LP.}
\label{tab:ablation300WLP}
\begin{adjustbox}{width=1\textwidth}
\begin{tabular}{c|c|l|l|l|l|l|l|l|c|l|l|l|l|l|l|l}
\hline
training set &
  \multicolumn{16}{c}{300W-LP} \\ \hline
testing set &
  \multicolumn{8}{c|}{AFLW2000} &
  \multicolumn{8}{c}{BIWI} \\ \hline
component 1 &
  \multicolumn{4}{c|}{-} &
  \multicolumn{4}{l|}{Attention mapping} &
  \multicolumn{4}{c|}{-} &
  \multicolumn{4}{l}{Attention mapping} \\ \hline
component 2 &
  \multicolumn{2}{c|}{-} &
  \multicolumn{2}{l|}{orthogonality} &
  \multicolumn{2}{c|}{-} &
  \multicolumn{2}{l|}{orthogonality} &
  \multicolumn{2}{c|}{-} &
  \multicolumn{2}{l|}{orthogonality} &
  \multicolumn{2}{c|}{-} &
  \multicolumn{2}{l}{orthogonality} \\ \hline
component 3 &
  - &
  \multicolumn{1}{c|}{Capsule} &
  - &
  \multicolumn{1}{c|}{Capsule} &
  \multicolumn{1}{c|}{-} &
  \multicolumn{1}{c|}{Capsule} &
  \multicolumn{1}{c|}{-} &
  \multicolumn{1}{c|}{Capsule} &
  \multicolumn{1}{c|}{-} &
  \multicolumn{1}{c|}{Capsule} &
  \multicolumn{1}{c|}{-} &
  \multicolumn{1}{c|}{Capsule} &
  \multicolumn{1}{c|}{-} &
  \multicolumn{1}{c|}{Capsule} &
  \multicolumn{1}{c|}{-} &
  \multicolumn{1}{c}{Capsule} \\ \hline
MAE &
  \multicolumn{1}{l|}{5.120} & 4.977
   & 4.979   %7.873
   & 4.951
   & 4.883
   & 4.740
   & 4.866
   & \textbf{4.669}
   & 4.390
   & 4.280
   & 4.298 %6.169
   & 4.165
   & 4.175
   & 4.022
   & 4.204
   & \textbf{3.972}
   \\ \hline
MAEV &
  \multicolumn{1}{l|}{6.487} & 6.181
   & 6.306 %10.063
   & 6.274
   & 6.167
   & 6.058
   & 6.157
   & \textbf{5.989}
   & 6.450
   & 6.360
   & 6.255 %8.705
   & 6.186
   & 6.256
   & 5.871
   & 6.280
   & \textbf{5.864}
   \\ \hline
\end{tabular}
\end{adjustbox}
\end{table*}

\begin{table*}[!htbp]
\centering
\caption{Ablation study for different feature mapping methods (with/ without attention mapping) and loss items (with/ without orthogonality loss) and capsule network (with/without capsule network). Trained on BIWI.}
\label{tab:ablationBIWI}
\begin{tabular}{l|c|l|l|l|c|l|l|l}
\hline
training set & \multicolumn{8}{c}{BIWI (train)}                                                    \\ \hline
testing set  & \multicolumn{8}{c}{BIWI (test)}                                                     \\ \hline
component 1  & \multicolumn{4}{c|}{-}                   & \multicolumn{4}{c}{Attention mapping}   \\ \hline
component 2 &
  \multicolumn{2}{c|}{-} &
  \multicolumn{2}{c|}{orthogonality} &
  \multicolumn{2}{c|}{-} &
  \multicolumn{2}{c}{orthogonality} \\ \hline
component 3 &
  - &
  \multicolumn{1}{c|}{Capsule} &
  \multicolumn{1}{c|}{-} &
  \multicolumn{1}{c|}{Capsule} &
  - &
  \multicolumn{1}{c|}{Capsule} &
  \multicolumn{1}{c|}{-} &
  \multicolumn{1}{c}{Capsule} \\ \hline
MAE          & \multicolumn{1}{l|}{3.422}  & 2.978 & 3.333  & 2.835  & \multicolumn{1}{l|}{2.840} & 2.826 & 2.843 &  \textbf{2.801}\\ \hline
MAEV         & \multicolumn{1}{l|}{4.920}  & 4.185  & 4.791 & 4.069 & \multicolumn{1}{l|}{4.162} & 4.142 & 4.118 &  \textbf{4.061}\\ \hline
\end{tabular}
\end{table*}

\begin{table*}[!htbp]
\centering
\caption{Ablation study for using a uniform prediction module.}
\label{tab:abaltionuniform}
\begin{tabular}{c|l|l|l|l|l|l}
\hline
training set & \multicolumn{4}{c|}{300W-LP}                              & \multicolumn{2}{c}{BIWI (train)} \\ \hline
testing set  & \multicolumn{2}{c|}{AFLW2000} & \multicolumn{2}{c|}{BIWI} & \multicolumn{2}{c}{BIWI (test)} \\ \hline
Uniform prediction & - & \multicolumn{1}{c|}{\checkmark} & - & \multicolumn{1}{c|}{\checkmark} & - & \multicolumn{1}{c}{\checkmark} \\ \hline
MAE          & \textbf{4.669} & 4.953 & \textbf{3.972} & 4.500 & \textbf{2.801} &  3.302 \\ \hline
MAEV         & \textbf{5.989} & 6.245 & \textbf{5.864} & 6.764  & \textbf{4.061}  &    4.862 \\ \hline
\end{tabular}
\end{table*}

%^^^^^^^^^^ Ablation Tables ^^^^^^^^^^%

Table~\ref{tab:aflw2000} and~\ref{tab:biwi} show the results of our proposed TriNet and other methods tested on AFLW2000 and BIWI datasets respectively. All of them are trained on the 300W-LP dataset. Since TriNet predicts three orthonormal vectors, its Euler angle results are obtained through the conversion from the rotation matrix constructed of these three vectors.

As the tables demonstrate, deep learning based landmark-free approaches (FSA-Net, Hopenet and TriNet) outperform landmark based methods (3DDFA and Dlib) on both AFLW2000 and BIWI datasets. In Table~\ref{tab:aflw2000}, we can find that if measured by MAE, FSA-Net surpasses the Hopenet by a large margin. However, their MAEV results are close. Even though Quatnet achieves the best MAE results, we are unable to replicate the MAEV results since it is not open-source. Meanwhile, as shown in Table~\ref{tab:biwi}, Our proposed method achieves the best result under both MAE and MAEV when tested on  BIWI dataset. 

Table \ref{tab:biwi3070} shows the experiment results of $3$-fold cross validation on BIWI dataset using different methods. In this scenario, we only compare our proposed method with other RGB-based ones. We compare both the MAE and MAEV results and our proposed TriNet achieves the best performance.

\subsection{Error Analysis}

We conduct the error analysis of three landmark-free methods (FSA-Net, Hopenet and TriNet) on AFLW2000 dataset. The Euler angles' range of $[-99^{\circ}, 99^{\circ}]$ are equally divided to intervals that span $33^{\circ}$. %We obtain both the MAE and MAEV on each interval. 
The results are shown in Fig.~\ref{fig:eulererroraflw2000} and Fig.~\ref{fig:vectorerroraflw2000}.

The first thing worth noting is that prediction error of MAEV increases much more slowly than MAE as absolute values of pose angles increase. MAE can achieve about $60^\circ$ for large pitch and roll angles while MAEV has only around $30^\circ$. This conforms to our findings in section 1 that MAE fails to measure performance at large pose angles. 

We use Fig.~\ref{fig:euler_failure_cases} to further illustrate the reason. Since gimbal lock causes ambiguity issue to Euler angles, many researchers limit the yaw angle in the range of $(-90^\circ, 90^\circ)$ to ensure the representation of rotation is unique. However, this brings in a new issue. Assume the rotation is in the order of pitch $(\gamma)$, yaw $(\beta)$ and roll $(\alpha)$ and denoted by $(\alpha, \beta, \gamma)$. As yaw exceeds the boundary $\pm 90^\circ$, it will cause significant change in pitch and roll angles. For example, assume a person with head pose of $(10^\circ, 89^\circ, 15^\circ)$. If he increases the yaw angle by $3^\circ$, this causes no observable difference in image. However, the annotation becomes $(-170^\circ, 88^\circ, -165^\circ)$ since yaw angle is not allowed to surpass $90^\circ$. This explains why similar profile images have very different Euler angle labels. As a result, Euler angle based network models hardly learn anything from such profile images. Fig.~\ref{fig:euler_failure_cases} verifies its validity. It shows the prediction results of different methods on profile images from AFLW2000. FSA-Net and Hopenet have very arbitrary results whereas our TriNet has accurate predictions. This figure also shows the problem of MAE. By comparing the MAE and MAEV results of TriNet (4th row), we can conclude that MAE cannot measure performance of networks on profile images.

The second noticeable thing is that the yaw angle error distribution is different from those of pitch and roll. As the absolute values of ground truth angles grow, the MAEV of yaw grows much more slowly compared with pitch and roll. We attribute this to their different distributions in the 300W-LP dataset.Data samples have their yaw angles evenly distributed across $[-99^{\circ}, +99^{\circ}]$ whereas $93\%$ pitch and $90\%$ roll angles concentrate in the range of $[-33^{\circ}, +33^{\circ}]$. Shortage of training data for large pitch and roll angles makes their performances worse than the yaw angle.

\subsection{Ablation Study}

In this section, we conduct ablation studies to analyze how each network component will affect the model performance on different testing sets. We include feature mapping methods, loss item, and capsule module as three testing components. Table \ref{tab:ablation300WLP} and \ref{tab:ablationBIWI} report both the MAE and MAEV results. We observe the best results on all the testing sets when combining all these three modules. %, showing that the fusion of different model variants enable to learn complementary feature information. 
In addition, in prediction module, we experiment the influence of uniform sampling. In other words, we use $(9,9,9)$ as the dimensions of three output vectors from capsule network instead of $(81, 27, 9)$. %we experiment to verify the influence of  sampling used in prediction module on the final results. 
Table~\ref{tab:abaltionuniform} shows non-uniform sampling can achieve better results.

%---------- Section 5 ----------%
\section{Conclusion}
In this paper, we put forward a new vector-based annotation and a new metric MAEV. They can solve the discontinuity issues caused by Euler angles. By the combination of new vector representation and our TriNet, we achieve state-of-the-art performance on the task of head pose estimation.

% In this paper, we show that neither Euler angles nor quaternions should be used for neural network training and that MAE is an inappropriate measurement for head pose estimation. Instead, we propose a new annotation method which uses three orthonormal vectors to represent rotation and a new evaluation criterion (MAEV) which measures angles between the ground truth and predicted vectors. %Based on that, a fine-grained architecture that utilizes multi-stage information is proposed. By introducing attention mechanism and feature grouping module, we 
%and achieves state-of-the-art results on both AFLW2000 and BIWI datasets. %This vector-based representation, even though we only demonstrate its effectiveness on head pose estimation, we believe it can be extended to other pose estimation problems as well.

{\small
\bibliographystyle{ieee_fullname}
\bibliography{egpaper}
}

\end{document}